\documentclass[final,10pt,times]{elsarticle}

\usepackage{amssymb}
\usepackage{amsmath}

\usepackage[ruled]{algorithm2e} 

\usepackage{algorithmic}

\usepackage{subfigure}
\usepackage{multirow}
\usepackage{multicol}
\usepackage{tablefootnote}
\usepackage{enumerate}
\usepackage{enumitem}
\usepackage{extarrows}

\usepackage{booktabs}
\usepackage{mathtools}
\usepackage{multirow}
\usepackage{subfigure}
\usepackage{caption}
\usepackage{subcaption}
\usepackage{graphicx}
\usepackage{url} 
\usepackage{pifont}
\usepackage{extarrows}
\usepackage{wrapfig}
\usepackage{floatrow}
\newfloatcommand{capbtabbox}{table}[][\FBwidth]
\journal{Pattern Recognition}

\begin{document}

\begin{frontmatter}



\title{Active Sampling for Node Attribute Completion on Graphs}


\author[label1]{Benyuan Liu} 
\author[label2]{Xu Chen}
\author[label1]{Yanfeng Wang}
\author[label1]{Ya Zhang}
\author[label3]{Zhi Cao}
\author[label4]{Ivor Tsang}

\affiliation[label1]{organization={Shanghai Jiao Tong University},
            }
\affiliation[label2]{{Pailitao, Alibaba Group},
            } 
            
\affiliation[label3]{organization={Academy of Broadcasting Planning, NRTA},
            }
\affiliation[label4]{{The Agency for Science, Technology and Research (A*STAR)},
            }

\begin{abstract}
Node attribute, a type of crucial information for graph analysis, may be partially or completely missing for certain nodes in real world applications. Restoring the missing attributes is expected to benefit downstream graph learning. Few attempts have been made on node attribute completion, but a novel framework called Structure-attribute Transformer (SAT) was recently proposed by using a decoupled scheme to leverage structures and attributes. SAT ignores the differences in contributing to the learning schedule and finding a practical way to model the different importance of nodes with observed attributes is challenging. This paper proposes a novel AcTive Sampling algorithm (ATS) to restore missing node attributes. The representativeness and uncertainty of each node's information are first measured based on graph structure, representation similarity and learning bias. To select nodes as train samples in the next optimization step, a weighting scheme controlled by Beta distribution is then introduced to linearly combine the two properties. 
Extensive experiments on four public benchmark datasets and two downstream tasks have shown the superiority of ATS in node attribute completion.
\end{abstract}




\end{frontmatter}



\section{Introduction}
\label{sec1}

Node attribute, a type of important information for graphs, plays an important role in many graph learning tasks, such as node classification~\cite{jin2021heterogeneous,xu2019adaptive} and community detection~\cite{sun2021graph,chen2017supervised}. The recent Graph Neural Network (GNN) enjoys boosted performance leveraging node attributes~\cite{defferrard2016convolutional,kipf2017semi,xu2018how,velickovic2018graph}. 
Despite of its indispensability, real-world graphs often are associated with missing node attributes for various reasons~\cite{chen2022learning}. For example,  in
citation graphs, key terms or detailed content of some papers may be inaccessible because of copyright protection. In social networks, profiles of some users may be unavailable due to privacy protection.
As a result, node attribute completion, which learns to restore the missing node attributes of a graph, has become an important research direction in graph learning community, given its expected benefits for downstream graph learning tasks.

So far, there are few attempts on the node attribute completion. 
The popular random walk based method~\cite{cui2018survey} and GNN framework are not designed for this task.
\cite{chen2022learning} propose a novel structure-attribute transformer (SAT) framework for node attribute completion, which leverages structures and attributes in a decoupled scheme and achieves the joint distribution modeling by matching the latent codes of structures and attributes. There are also several methods to deal with the attribute missing graph such as WGNN~\cite{chen2021wasserstein}, HGNN-AC~\cite{jin2021heterogeneous}, PaGNNs~\cite{jiang2020incomplete}, Amer~\cite{amer}. They apply different attribute learning methods. WGNN encodes the attribute matrix into the Wasserstein space and reconstructs to Euclidean representation. HGNN-AC obtains node topological embeddings and complete the missing attributes by using topological relationship. PaGNNs propose partial aggregation functions and can be performed directly on an incomplete graph. Generally they incorporate the structural information by neighborhood aggregation. Amer is a unified model to combine the processes of completing missing attributes and learning embedding, but they mainly focus on coarse-grained task such as node classification.


Although SAT has shown great promise on node attribute completion compared with other methods, it treats the nodes with observed attributes equally and ignores their difference in contributing to the learning schedule. 
Importance re-weighting~\cite{wang2017robust,fang2020rethinking,byrd2019effect} on the optimization objective seems to be a potential solution.
However, the information of nodes is influenced mutually and has more complex patterns, thus making the importance distribution implicit, intractable and rather complicated. 
It's challenging to find a practical way to model contribution of the nodes with observed attributes.

This paper proposes \textbf{A}c\textbf{T}ive \textbf{S}ampling (ATS) to better leverage the partial nodes with observed attributes. 
In particular, ATS measures the \textbf{\textit{representativeness}} and \textbf{\textit{uncertainty}} of each node by considering the graph structures, representation similarity and learning bias.
Based on the measurement, it then adaptively and gradually selects nodes from the candidate set to the train set after each training epoch, encouraging the model to consider the node's importance in learning. 
When determining how to merge the above metrics, it is ideal for the learning process to prioritize nodes of higher representativeness at the early stage when the model is unsteady and uncertainty metric is not sufficiently dependable. As the model becomes more trustworthy, it may consider the uncertainty at the later stage. Thereby, we propose a Beta distribution controlled weighting scheme to exert adaptive learning weights on representativeness and uncertainty. In this way, these two metrics are linearly combined as the final score to automatically determine which nodes are selected into the train set in next optimization epoch. ATS can be combined with existing node attribute completion models (e.g. SAT) and learned in an iterative manner until the model converges. 
Extensive experiments on four public benchmarks have shown that ATS can help the node attribute completion model reach a better optimum, and restore higher-quality node attributes that benefit downstream node classification and profiling tasks. In addition, ATS can be flexibly applied to different primary models.
The contributions of the paper are as summarized follows:
\begin{itemize}
    \item To better leverage the partial nodes with observed attributes for node attribute completion, active sampling is proposed to adaptively and gradually select samples into the train set in each optimization epoch.
    \item We propose  to measure the importance of nodes with \textbf{\textit{representativeness}} and \textbf{\textit{uncertainty}}. 
    \item A Beta distribution controlled weighting scheme is proposed to combine \textbf{\textit{representativeness}} and \textbf{\textit{uncertainty}} as the final score function. 
\end{itemize}

\section{Related work}
\subsection{Deep Graph Learning}
\label{subsec1}

With the development of deep representation learning in the Euclidean vision domain~\cite{10.1155/2018/7068349}, researchers have studied a lot of deep learning methods on the non-Euclidean graphs~\cite{9039675}.
Random walk based methods 
can learn node embeddings by random walks
, which only considers the structural information and cannot generalize to new graphs. To tackle this problem, the attributed random walk based methods (e.g.GraphRNA~\cite{huang2019graph}) apply random walks on both structures and attributes.
These random walk based methods are useful, but they demand hardly-acquired high-quality random walks
to guarantee good performance.
Graph Neural Network (GNN)~\cite{scarselli2008graph,defferrard2016convolutional,kipf2017semi} realizes "graph-in, graph-out" that transforms the embeddings of node attributes while maintaining the connectivity~\cite{sanchez-lengeling2021a}. GNN performs a message passing scheme, which is reminiscent of standard convolution as in Graph Convolutional Networks (GCN)~\cite{kipf2017semi}. GNN can infer the distribution of nodes based on node attributes and edges and achieve impressive results on graph-related tasks. 
There are also numerous creative modifications in GNN such as GAT~\cite{velickovic2018graph} and GraphSAGE~\cite{hamilton2017inductive}.

In last few years, more works have emphasized the importance of node attributes in graph-related downstream tasks.
Both SEAL~\cite{pan2022neural} and WalkPool ~\cite{zhang2018link} encode node representations with node attributes to achieve superior link prediction performance. In most real-world applications, attributes of some nodes may be inaccessible, so the node attribute completion task appears. 
Recent SAT~\cite{chen2022learning} assumes a shared-latent space assumption on graphs and proposes a novel GNN-based distribution matching model. It decouples structures and attributes and simultaneously matches the distribution of respective latent vectors. WGNN developed by ~\cite{chen2021wasserstein} learns node representations in Wasserstein space without any imputation. ~\cite{jin2021heterogeneous} propose HGNN-AC to learn topological embedding and attribute completion with weighted aggregation. PaGNNs~\cite{jiang2020incomplete} restores the missing attributes based on a partial message propagation scheme. These methods implement several aggregation methods to incorporate the structural information, while SAT's shared latent space assumption matches the joint distribution of structure and attributes by two distinct encoders. HGCA~\cite{HGCA} is proposed for heterogeneous network with missing attributes. Feature Propagation~\cite{rossi2021fp} and Amer~\cite{amer} can resolve high missing rate of the node attributes, but they generally focus on coarse-grained task such as node classification. Note that they all do not model the different contributions of nodes with observed attributes in learning, while the proposed ATS can help them achieve this goal.

\subsection{Active Sampling on Graphs}

Active learning assists the model to achieve as better performance as possible while labeling as few samples as possible~\cite{ren2021survey}. It's usually combined with deep learning primary model to select the most influential samples from unlabeled dataset and then label them for training to reduce the annotation cost~\cite{yoo2019learning}. 
There are also some works of active learning on graph data. 
Early works~\cite{gadde2014active,gu2013selective,ji2012variance} mainly take graph structures into consideration and design the query strategy regardless of node attributes. With the development of deep learning, many active learning algorithms are designed based on GNN. The query strategy of AGE~\cite{cai2017active} measures the amount of the information contained in different nodes to select the most informative candidate node. Similar to AGE, ANRMAB~\cite{gao2018active} adopts the weighted sum of different heuristics, but it adjusts the weights based on a multi-armed bandit framework. 
~\cite{caramalau2021sequential} discuss two novel sampling methods: UncertainGCN and CoreGCN, which are based on uncertainty sampling and CoreSet~\cite{sener2017active}, respectively.

Nevertheless, most of today's popular active sampling algorithms on graph aim to resolve the node classification task and focus on how to reduce the annotation cost. For this node attribute completion task, since the attribute-observed nodes are limited and the dimension of node attributes is much higher than node classes, we demand a more advanced active sampling algorithm to help the primary model
utilize the attribute-observed nodes more effectively and learn the complicated attribute distribution better. In addition, the current query strategies measure the uncertainty by an unsupervised manner, but we propose a supervised one to make the sampling closer to the node attribute completion target.


\section{Problem Formulation}
For node attribute completion task, we denote $\mathcal{G} = (\mathcal{V},A,X)$ as a graph with node set $\mathcal{V} = \{v_1,v_2,\ldots, v_N\}$, the adjacency matrix $A\in R^{N\times N}$ and the node attribute matrix $X\in R^{N\times F}$. $\mathcal{V}^o = \{v_1^o, v_2^o, ..., v_{N_o}^o\}$ is the set of attribute-observed nodes. The attribute information of $\mathcal{V}^o$ is $X^o = \{x_1^o, x_2^o, ..., x_{N_o}^o\}$ and the structural information of $\mathcal{V}^o$ is $A^o = \{a_1^o, a_2^o, ..., a_{N_o}^o\}$. $\mathcal{V}^u = \{v_1^u, v_2^u, ..., v_{N_u}^u\}$ is the set of attribute-missing nodes. The attribute information of $\mathcal{V}^u$ is $X^u = \{x_1^u, x_2^u, ..., x_{N_u}^u\}$ and the structural information of $\mathcal{V}^u$ is $A^u = \{a_1^u, a_2^u, ..., a_{N_u}^u\}$. More specifically, we have $\mathcal{V} = \mathcal{V}^u \cup \mathcal{V}^o$, $\mathcal{V}^u \cap \mathcal{V}^o = \emptyset$, and $N = N_o + N_u$. 
In recent proposed works~\cite{chen2022learning,chen2021wasserstein,jiang2020incomplete}, learning the latent representations of attribute-missing nodes $\mathcal{V}^{u}$ based on the available structures $A$ together with observed node attributes $X^o$, and then translating the latent representations to missing node attributes $X^{u}$ is a commonly recognized way. The difference among these works is the technique of learning latent representations and translating features. 
Among existing models, SAT performs well and has open source codes, so we will refer to SAT as a primary base model to verify the proposed algorithm in later experiments.

In our active sampling algorithm, we denote the total training set as $T$, in which the node attributes are known. The current training set of primary base model is $T^L$ and the set containing candidate nodes is denoted as $T^U$. We have $T = T^L \cup T^U$. We design a reasonable sampling strategy named ATS which iteratively transfers the most suitable candidate nodes from $T^U$ to $T^L$ to boost the training effectiveness of primary model until $T^U = \emptyset$ and the model converges.

\section{Method}
We design query strategy of the ATS by measuring the \textbf{representativeness} and \textbf{uncertainty} of the candidate nodes. Then we combine the uncertainties and representativeness scores as the final score using an adaptive weighting scheme and select the nodes with the highest scores for the next learning epoch. We will explain these in more detail in the following parts.

\subsection{Query Strategy}
\label{sec: query strategy}
\textbf{Representativeness}: The major and typical patterns among the nodes are vital for the model to converge to the right direction.
In this section, we introduce the concept of representativeness as a sampling metric.
This metric is composed of two parts: 
1) information density $\phi_{density}$ and 2) structural centrality $\phi_{centrality}$.
The former mainly focuses on measuring the similarity between the corresponding latent vectors of attributes and structures. The latter indicates how closely a node is connected to its neighbours on graph. In other words, the information density is inspired by the good representation learning ability of primary model and the structural centrality is natural to mine the information on the graph structures. These two aspects offer us a comprehensive analysis of the representativeness in both implicit and explicit ways.

We first focus on the information density. 
In particular, we measure the node similarities through the latent features learned by the primary model. If there is a dense distribution of representation vectors in a local region of the latent space, the corresponding nodes will have more similar features and this region will contain further mainstream information, so we expect to train these more representative nodes in priority. 
In the case of SAT, there are attribute embeddings of attribute-observed nodes and structure embeddings of all nodes in SAT. ATS only uses the structure embeddings $z_{a_i}$ to calculate the $\phi_{density}$ as shown in Eq.~\ref{density} since we believe the structural representations have more information of density on graphs.
In order to find the central node located in high-density region, we employ the K-means algorithm in the latent space and calculate the Euclidean distance between each node and its clustering center. Given $d$ as the metric of Euclidean distance in $l_{2}$-norm and $C_{z_{a_i}}$ as the clustering center of $z_{a_i}$ in latent space, the formulation of $\phi_{density}$ is written as:
\begin{equation}
\label{density}
    \phi_{density}(v_i) = \frac{1}{1+d(z_{a_i}, C_{z_{a_i}})}, v_i \in T^U
\end{equation}
The larger the $\phi_{density}$ is, the more representative the node is, 
and the node contains more representative features that are worthy of the model's attention.

Besides the feature analysis in latent space, the node representativeness can also be inferred from the explicit graph structures. We can study the connections between nodes and develop a metric to calculate the node centrality based on the structural information. Intuitively, the centrality can have a positive correlation with the number of neighbours. 
At the early stage of training, if we can focus on these nodes, the model will learn the approximate distribution of the data faster and reduce the influence caused by the noisy ones. PageRank~\cite{page1999pagerank} algorithm is an effective random-walk method to acquire the visiting probabilities of nodes.The higher score signifies the higher visiting probabilities, which means that nodes have relatively more neighbors and then contain more structural information. We find that PageRank is the most suitable one because it has well ability of representing centrality~\cite{cai2017active}.

We also utilize the PageRank score as the structural centrality $\phi_{centrality}$, which is shown as below:
\begin{equation}
    \phi_{centrality}(v_i) = \rho \sum_{j}A_{ij}\frac{\phi_{centrality}(v_j)}{\sum_k A_{jk}} + \frac{1-\rho}{N^U}, v_i \in T^U
\end{equation}
where $N^U$ is the number of nodes in $T^U$, $\rho$ is the damping parameter.
The larger $\phi_{centrality}$ is, the more representative the node is,
and the node is more closely associated with its neighbours.

\textbf{Uncertainty}: Uncertainty reflects the learning state of the current model towards the nodes. When the model is reliable, it's reasonable to pay more attention on the nodes that have not been sufficiently learned. Uncertainty is a commonly-used query criterion in active learning. However, as mentioned before, the uncertainty in other sampling algorithms~\cite{cai2017active,caramalau2021sequential,zhang2022information} usually works for node classifications and is designed in an unsupervised manner to reduce the annotation cost. 
In this paper, we directly refer to the training loss of primary model as the uncertainty metric. For the node attribute completion task, in order to know the training status of the model more accurately, we consider the observed attributes and structures as supervision, and use the learning loss of primary model as the uncertainty metric, noted as $\phi_{entropy}(v_i)$.
\begin{equation}
    \phi_{entropy}(v_i) = \mathcal{L}(v_i), v_i \in T^U
\end{equation}
 where $\mathcal{L}$ denotes the loss of primary model. We can input the attributes of candidate nodes and the corresponding graph structures into specific primary model, and then obtain their loss values. 
The larger $\phi_{entropy}(v_i)$ is, the more uncertainty of node $v_i$ has. From the perspective of information theory, nodes with greater uncertainty contain more information. Sampling these nodes can help the model get the information that has not been learned in previous training, thus helping the training.

\subsection{Score function and Beta distribution controlled weighting scheme}
We have presented three metrics of our query strategy. 
Then, a question arises: How to combine these metrics to score each node? Combing the metrics with a weighted sum is a possible solution but still faces great difficulties.
First, the values of different metrics are incomparable because of the distinct dimensional units. Second, the different metrics may take different effects at different learning stages.
To solve these, we introduce a percentile evaluation and design a Beta-distribution controlled re-weighting scheme 
to exert the power of each metric, since Beta distribution is a suitable model for the random behavior of percentages and proportions~\cite{gupta2004handbook}.

Denote $\mathcal{P}_\phi(v_i, T^U)$ as the percentage of the candidate nodes in $T^U$ which have smaller values than the node $v_i$ with metric $\phi$. For example, if there are 5 candidate nodes and the scores of one metric is $[1, 2, 3, 4, 5]$, the percentile of the corresponding 5 nodes will be $[0, 0.2, 0.4, 0.6, 0.8]$. We apply the percentile to three metrics and define the final score function of ATS as:
\begin{align}
    S(v_i) = &\alpha\cdot\mathcal{P}_{entropy}(v_i, T^U) + \beta\cdot\mathcal{P}_{density}(v_i, T^U)\nonumber\\
    &+ \gamma\cdot\mathcal{P}_{centrality}(v_i, T^U)
\end{align}
where $\alpha+\beta+\gamma = 1$. At the sampling stage, ATS will select one or several nodes with the largest $S$ and add them to the training set $T^L$ for the next training epoch.

\begin{algorithm}[!htp]
  \caption{ATS algorithm}
  \label{algo:algorithm}
  \small
  \textbf{Input}: Parameters of the primary base model--SAT, $T^U$, $T^L$
  initialization of $T^L$ and hyper-parameters;\\
  \begin{algorithmic}[1]
  \WHILE{$n_{e} < total\_epoch$}
  \STATE $loss \leftarrow SAT(T^L)$;\COMMENT{//Training stage}
  \\
  \STATE $loss.backward()$;
  \STATE $update(SAT.params)$;\COMMENT{//Update SAT}
  \\
  \IF{$\# T^U > 0$} 
  \STATE\COMMENT{//SAT returns loss values and latent representations $z_a$}
    $z_a, \phi_{entropy} \leftarrow SAT(T^U)$;\\
    $\phi_{density} \leftarrow getDensity(z_a)$;\\
    $\phi_{centrality} \leftarrow getCentrality(G)$;
  \STATE $\gamma \leftarrow Beta(1, n_t)$;
    $\alpha, \beta \leftarrow \frac{1-\gamma}{2}$;
  \STATE $S \leftarrow \alpha\cdot\mathcal{P}_{entropy} + \beta\cdot\mathcal{P}_{density} + \gamma\cdot\mathcal{P}_{centrality}$;
    \\
  \STATE $T^S \leftarrow activeSample(S,T^U)$;\\ \COMMENT {//select the node with the highest score}
  \STATE $T^L \leftarrow T^L \cup T^S$; \COMMENT{//renew the training set of SAT}
  \STATE $T^U \leftarrow T^U\setminus T^S$;\COMMENT{//renew the candidate set}
  \ENDIF
  \ENDWHILE
  \end{algorithmic}
\end{algorithm}

Further, it is worth noting that the uncertainty and the information density are determined by the training result returned from SAT. At an early training stage, the model is unstable and the returned training result may not be quite reliable. A sampling process based on inaccurate model-returned results may lead to undesirable results. Hence, we set the weights to time-sensitive ones. The structure-related weight $\gamma$ is more credible so it can be larger initially. As the training epoch increases, the model can pay more attention to $\phi_{entropy}$ and $\phi_{density}$, while the weight $\gamma$ will decrease gradually. We formalize this by sampling $\gamma$ from a Beta distribution, of which the expectation becomes smaller with the increase of training epoch. The weighting values are defined as:
\begin{equation}
    \label{gamma}
    \begin{cases}
    \gamma = 1, & \quad n_e\leq threshold \\
     \gamma \sim Beta(1, n_t), & \quad n_t = \frac{n_{e}-threshold}{\epsilon}~~and~~\alpha = \beta = \frac{1 - \gamma}{2}.
     \end{cases}
\end{equation}


where $n_{t}$ is one of the determinants in Beta distribution; $\epsilon$ is used to control the expectation of $\gamma$; $n_{e}$ denotes the current number of epochs. We obtain the expectation by calculating the average value of 10,000 random samples. 

\subsection{Iterative training and Implementation}
In general, our method consists of two stages: one is primary base model SAT, responsible for the training stage; the other is ATS, responsible for the sampling stage. 
Before the training, we divide total training set $T$ into $T^U$ and $T^L$. We randomly sample $1\%$ of the nodes in $T$ as the initial nodes of $T^L$ and the rest composes $T^U$. SAT will be trained on the changeable $T^L$.
Once SAT accomplishes a single training epoch, ATS starts the sampling process. We sample the most representative and informative candidate nodes from $T^U$ according to the query strategy. These selected nodes are added to $T^L$ and removed from $T^U$. Then SAT will be trained on the renewed $T^L$ at next epoch. The training stage and the sampling stage alter iteratively until $T^U$ is null. Finally ATS is terminated and SAT will continue training to convergence.
We clarify the learning process in Algorithm~\ref{algo:algorithm}.

\section{Experiments and Analysis}
\subsection{Datasets}
We utilize 4 public benchmarks whose node attributes are categorical vectors instead of numeric ones. For numeric attributes, SAT implements auto-encoding after normalization whose optimization is not stable according to our experiments. 
The information of used datasets is as follows: \textbf{1) Cora}. Cora \cite{mccallum2000automating} is a citation graph with 2,708 papers as nodes and 10,556 citation links as edges. Each node has a multi-hot attribute vector with 1,433 dimensions. The attribute vectors consist of different word tokens to determine whether they appear or not.
\textbf{2) Citeseer}. Citeseer~\cite{sen2008collective} is another citation graph which is larger than Cora. It contains 3,327 nodes and 9,228 edges. Like Cora, each node has a multi-hot attribute vector with 3,703 dimensions.
\textbf{3) Amazon-Computer} and \textbf{4) Amazon-Photo}. These two datasets are generated from Amazon co-purchase graph. The node represents the item and the edge represents the two items are usually purchased at the same time. The node attribute is a multi-hot vector with the set of words involved in the item description. Amazon-Computer~\cite{shchur2018pitfalls} has 13,752 items and 245,861 edges. Amazon-Photo~\cite{shchur2018pitfalls} has 7,650 nodes and 119,081 edges.

\subsection{Experimental setup}
\textbf{Baselines}: 
We compare ATS with other baselines introduced in ~\cite{chen2022learning}: NeighAggre~\cite{csimcsek2008navigating}, VAE~\cite{kingma2013auto}, GCN~\cite{kipf2017semi}, GraphSage~\cite{hamilton2017inductive}, GAT~\cite{velickovic2018graph}, Hers~\cite{hu2019hers}, GraphRNA~\cite{huang2019graph}, ARWMF~\cite{chen2019}, PaGNN~\cite{jiang2020incomplete} and original SAT~\cite{chen2022learning}. Details about how they work on node attribute completion are illustrated in Appendix~\ref{appendix:baseline_details}.

\textbf{Parameters setting}: In the experiment, we randomly sample 40\% nodes with attributes as training data, 10\% nodes as validation data and the rest as test data. The attributes of validation and test nodes are unobserved in training.
For the baselines, the parameters setting and the experiment results refer to \cite{chen2022learning}. For our ATS method, the SAT's setting remains the same, such as $\lambda_c$. We mainly have two hyper-parameters: $\epsilon$ in the weighting scheme and cluster numbers in the estimation of density $\phi_{density}$. Considering the objective of the Beta distribution weighting scheme, $\epsilon$ should be larger than the total sampling times. Hence in Cora and Citeseer, we set $\epsilon = 1500$ and when it comes to Amazon\_Photo and Amazon-Computer, we set $\epsilon = 2000$. In addition, we set the cluster number as 10, 12, 15, 20 for Cora, Citeseer, Amazon-Photo and Amazon-Computer. The \textit{threshold} for Cora is 300 and the rest is 0.

\textbf{Evaluation metrics}: In node attribute completion, the restored attributes can provide side information for nodes and benefit downstream tasks. By following SAT~\cite{chen2022learning}, we study the effect of ATS on two downstream tasks including node classification task in the node level and profiling task in the attribute level. In node classification, we train an extra classifier based on the recovered attributes of test nodes to evaluate whether the restored attributes can serve as one kind of data augmentation and benefit the classification model. In profiling, we aim to predict the possible profile (e.g. word tokens in Cora) in each attribute dimension and evaluate the recall and ranking quality.

\subsection{Overall Comparison}
\subsubsection{Node Classification}
Classification is an effective downstream task to test the quality of the recovered attributes. In node classification task, the nodes with restored attributes are split into 80\% training data and 20\% test data. Then we conduct five-fold cross-validation in 10 times and take the average results of evaluation metrics as the model performance. We use two supervised classifiers: MLP and GCN. The MLP classifier is composed by two fully-connected layers, which classifies the nodes based on attributes. The GCN classifier is an end-to-end graph representation learning model, which can learn the structure and attributes simultaneously. Results are shown in Table~\ref{tab:classification}.

According to the results of "X" row where only node attributes are used, most of the optimized models with our proposed ATS algorithm achieve obvious improvement compared with original models, especially those models that are not designed specifically for attribute missing graph such as GCN, GAT etc. 
Our ATS can also adapt to SAT with different GNN backbones (e.g. GCN and GAT) and achieve higher classification accuracy than the original models. 
For the results of "A+X" row where both structures and node attributes are used by a GCN classifier,
the performance of our ATS combined with other traditional models (e.g. GCN, GAT, GraphSage) achieves an increase of more than 15\% compared to that of original models on all datasets, because ATS contains the density metric and can help the model better learn the inner semantic structures. Furthermore, we can observe the increment on other newly proposed models, which reflects the generalizability of our method.

\begin{table}
    \vspace{-5pt}
    \small
    \captionsetup{font=small}
    \captionsetup{skip=0.1\textwidth}
    \caption{Node classification of the node-level evaluation for node attribute completion.
    "X" indicates the MLP classifier that only considers the node attributes. "A+X" indicates the GCN classifier that considers both the structures and node attributes. Values in parentheses signify the performance increment versus respective original model.}
    \label{tab:classification}
    \centering
\renewcommand{\arraystretch}{1.0}
 \setlength{\tabcolsep}{1mm}{ 
  \scalebox{0.69}{
    \begin{tabular}{c|cccc|cccc}
    \toprule
         & \multicolumn{4}{c}{X} & \multicolumn{4}{|c}{A+X} \\
         \cmidrule(r){2-9}
         \multirow{2}*{Method} & \multirow{2}*{Cora} & \multirow{2}*{Citeseer} & Amazon & Amazon & \multirow{2}*{Cora} & \multirow{2}*{Citeseer} & Amazon & Amazon  \\ 
         &  &  & Computer & Photo &  &  & Computer & Photo \\ 
         \midrule
        NeighAggre & 0.6248 & 0.5539 & 0.8365 & 0.8846 & 0.6494 & 0.5413 & 0.8715 & 0.901\\
        VAE & 0.2826 & 0.2551 & 0.3747 & 0.2598 & 0.3011 & 0.2663 & 0.4023 & 0.3781\\
        GCN & 0.3943 & 0.3768 & 0.3660 & 0.2683 & 0.4387 & 0.4079 & 0.3974 & 0.3656\\
        GraphSage & 0.4852 & 0.3933 & 0.3747 & 0.2598 & 0.5779 & 0.4278 & 0.4019 & 0.3784\\
        GAT & 0.4143 & 0.2129 & 0.3747 & 0.2598 & 0.4525 & 0.2688 & 0.4034 & 0.3789\\
        Hers & 0.3046 & 0.2585 & 0.3747 & 0.2598 & 0.3405 & 0.3229 & 0.4025 & 0.3794\\
        GraphRNA & 0.7581 & 0.6320 & 0.6968 & 0.8407 & 0.8198 & 0.6394 & 0.8650 & 0.9207\\
        ARWMF & 0.7769 & 0.2267 & 0.5608 & 0.4675 & 0.8025 & 0.2764 & 0.7400 & 0.6146\\
        PaGNN & 0.7638 & 0.6258 & 0.8331 & \textbf{0.9092} & 0.8148 & 0.6569 & \textbf{0.8832} & 0.9248\\
        SAT(GCN) & 0.7644 & 0.6010 & 0.7410 & 0.8762 & 0.8327 & 0.6599 & 0.8519 & 0.9163\\
        SAT(GAT) & 0.7937 & 0.6475 & 0.8201 & 0.8976 & \textbf{0.8579} & 0.6767 & 0.8766 & 0.9260\\ \cmidrule(r){1-9}
        \multirow{2}*{ATS+GCN} & 0.5015 & 0.5259 & 0.3861 & 0.3602 & 0.5742 & 0.5522 & 0.4795 & 0.5552\\
        & (+27.19\%) & (+39.57\%) & (+5.49\%) & (+34.25\%) & (+30.89\%) & (+35.38\%) & (+20.66\%) & (+51.86\%)\\
         \multirow{2}*{ATS+GAT} & 0.6056 & 0.5403 & 0.3959 & 0.3736 & 0.6795 & 0.5745 & 0.4843 & 0.5274\\
        & (+46.17\%) & (+153.78\%) & (+5.66\%) & (+43.80\%) & (+50.17\%) & (+113.73\%) & (+20.05\%) & (+39.19\%)\\
         \multirow{2}*{ATS+GraphSage} & 0.6148 & 0.4676 & 0.4028 & 0.4014 & 0.7292 & 0.5072 & 0.4741 & 0.6176\\
        & (+26.71\%) & (+18.89\%) & (+7.49\%) & (+54.50\%) & (+26.18\%) & (+18.56\%) & (+17.96\%) & (+63.21\%)\\
        \multirow{2}*{ATS+PaGNN} & 0.7701 & 0.6324 & \textbf{0.8419} & 0.9040 & 0.8197 & 0.6575 & \textbf{0.8832} & 0.9247\\
        & (+0.82\%) & (+1.05\%) & (+1.05\%) & (-0.57\%) & (+0.6\%) & (+0.09\%) & (+0.00\%) & (-0.01\%)\\
        \multirow{2}*{ATS+SAT(GCN)} & 0.7850 & 0.6370 & 0.8198 & 0.8827 & 0.8366 & 0.6750 & 0.8752 & 0.9181\\
        & (+2.69\%) & (+5.99\%) & (+10.63\%) & (+0.74\%) & (+0.47\%) & (+2.28\%) & (+2.74\%) & (+0.20\%)\\
        \multirow{2}*{ATS+SAT(GAT)} & \textbf{0.8065} & \textbf{0.6662} & 0.8402 & 0.9065 & 0.8573 & \textbf{0.6851} & 0.8822 & \textbf{0.9273}\\
        & (+1.61\%) & (+2.89\%) & (+2.45\%) & (+0.99\%) & (-0.07\%) & (+1.24\%) & (+0.64\%) & (+0.14\%)\\
        \bottomrule
    \end{tabular}
}}
\end{table}

\begin{table*}[!htp]
    \caption{Profiling of the attribute-level evaluation for node attribute completion.}
     \label{tab:profiling}
    \centering
    \normalsize
\renewcommand{\arraystretch}{0.7}
 \setlength{\tabcolsep}{1.2mm}{ 
  \scalebox{0.65}{
    \begin{tabular}[c]{c|c|cccccc}
    \toprule
    & Method & Recall@10 & Recall@20 & Recall@50 & NDCG@10 & NDCG@20 & NDCG@50 \\ \midrule
    \multirow{12}*{Cora}
         & GCN & 0.1271 & 0.1772 & 0.2962 & 0.1736 & 0.2076 & 0.2702  \\
         & GraphSage & 0.1234 & 0.1739 & 0.2842 & 0.1700 & 0.2041 & 0.2619 \\
         & GAT & 0.1350 & 0.1812 & 0.2972 & 0.1791 & 0.2099 & 0.2711 \\
         & PaGNN & 0.1482 & 0.2244 & 0.3516 & 0.2079 & 0.2590 & 0.3273 \\
         & SAT(GCN) & 0.1508 & 0.2182 & 0.3429 & 0.2112 & 0.2546 & 0.3212 \\
         & SAT(GAT) & 0.1653 & 0.2345 & 0.3612 & 0.2250 & 0.2723 & 0.3394 \\ \cmidrule(r){2-8}
         & ATS+GCN & 0.1285 & 0.1839 & 0.3066 & 0.1770 & 0.2144 & 0.2788 \\
         & ATS+GraphSage & 0.1214 & 0.1738 & 0.2862 & 0.1665 & 0.2014 & 0.2607 \\
         & ATS+GAT & 0.1296 & 0.1869 & 0.2994 & 0.1802 & 0.2189 & 0.2783 \\
         & ATS+PaGNN & 0.1511 & 0.2271 & 0.3545 & 0.2070 & 0.2581 & 0.3261 \\
         & ATS+SAT(GCN) & 0.1553 & 0.2232 & 0.3469 & 0.2135 & 0.2591 & 0.3250 \\
         & ATS+SAT(GAT) & \textbf{0.1671} & \textbf{0.2379} & \textbf{0.3616} & \textbf{0.2271} & \textbf{0.2745} & \textbf{0.3403} \\
    \hline
    \multirow{12}*{Citeseer}
    & GCN & 0.0620 & 0.1097 & 0.2052 & 0.1026 & 0.1423 & 0.2049 \\
    & GraphSage & 0.0559 & 0.0998 & 0.1944 & 0.0881 & 0.1247 & 0.1866 \\
    & GAT & 0.0561 & 0.1012 & 0.1957 & 0.0878 & 0.1253 & 0.1872 \\
    & PaGNN & 0.0799 & 0.1349 & 0.2514 & 0.1328 & 0.1786 & 0.2550 \\
     & SAT(GCN) & 0.0764 & 0.1280 & 0.2377 & 0.1298 & 0.1729 & 0.2447 \\
     & SAT(GAT) & 0.0811 & 0.1349 & 0.2431 & 0.1385 & 0.1834 & 0.2545 \\ \cmidrule(r){2-8}
     & ATS+GCN & 0.0666 & 0.1142 & 0.2107 & 0.1107 & 0.1501 & 0.2133 \\
     & ATS+GraphSage & 0.0587 & 0.1043 & 0.1921 & 0.0986 & 0.1367 & 0.1942 \\
     & ATS+GAT & 0.0659 & 0.1132 & 0.2131 & 0.1097 & 0.1491 & 0.2144 \\
     & ATS+PaGNN & 0.0800 & 0.1341 & 0.2502 & 0.1329 & 0.1781 & 0.2544 \\
     & ATS+SAT(GCN) & 0.0854 & 0.1400 & 0.2580 & 0.1441 & 0.1896 & 0.2672 \\
     & ATS+SAT(GAT) & \textbf{0.0917} & \textbf{0.1487} & \textbf{0.2635} & \textbf{0.1558} & \textbf{0.2037} & \textbf{0.2791} \\
     \hline
     \multirow{12}*{Amazon-Computer}
     & GCN & 0.0273 & 0.0533 & 0.1275 & 0.0671 & 0.1027 & 0.1824 \\
    & GraphSage & 0.0269 & 0.0528 & 0.1278 & 0.0664 & 0.1020 & 0.1822 \\
    & GAT & 0.0271 & 0.0530 & 0.1278 & 0.0673 & 0.1028 & 0.1830 \\
    & PaGNN & 0.0423 & 0.0749 & 0.1577 & 0.1039 & 0.1472 & 0.2354 \\
     & SAT(GCN) & 0.0391 & 0.0703 & 0.1514 & 0.0963 & 0.1379 & 0.2243 \\
     & SAT(GAT) & 0.0421 & 0.0746 & 0.1577 & 0.1030 & 0.1463 & 0.2346 \\ \cmidrule(r){2-8}
     & ATS+GCN & 0.0295 & 0.0555 & 0.1302 & 0.0717 & 0.1073 & 0.1874 \\
     & ATS+GraphSage & 0.0268 & 0.0528 & 0.1236 & 0.0666 & 0.1020 & 0.1784 \\
     & ATS+GAT & 0.0294 & 0.0556 & 0.1308 & 0.0719 & 0.1077 & 0.1883 \\
      & ATS+PaGNN & 0.0435 & 0.0765 & 0.1591 & 0.1059 & 0.1498 & 0.2376 \\
     & ATS+SAT(GCN) & 0.0421 & 0.0746 & 0.1575 & 0.1032 & 0.1464 & 0.2347 \\
     & ATS+SAT(GAT) & \textbf{0.0440} & \textbf{0.0775} & \textbf{0.1617} & \textbf{0.1074} & \textbf{0.1519} & \textbf{0.2412} \\
     \hline
      \multirow{12}*{Amazon-Photo}
      & GCN & 0.0294 & 0.0573 & 0.1324 & 0.0705 & 0.1082 & 0.1893 \\
    & GraphSage & 0.0295 & 0.0562 & 0.1322 & 0.0712 & 0.1079 & 0.1896 \\
    & GAT & 0.0294 & 0.0573 & 0.1324 & 0.0705 & 0.1083 & 0.1892 \\
    & PaGNN & 0.0433 & 0.0776 & 0.1647 & 0.1055 & 0.1510 & 0.2431 \\
     & SAT(GCN) & 0.0410 & 0.0743 & 0.1597 & 0.1006 & 0.1450 & 0.2359 \\
     & SAT(GAT) & 0.0427 & 0.0765 & 0.1635 & 0.1047 & 0.1498 & 0.2421 \\ \cmidrule(r){2-8}
     & ATS+GCN & 0.0310 & 0.0580 & 0.1336 & 0.0757 & 0.1125 & 0.1937 \\
     & ATS+GraphSage & 0.0300 & 0.0572 & 0.1324 & 0.0732 & 0.1101 & 0.1911 \\
     & ATS+GAT & 0.0307 & 0.0576 & 0.1342 & 0.0754 & 0.1121 & 0.1941 \\
     & ATS+PaGNN & 0.0435 & 0.0776 & 0.1648 & 0.1059 & 0.1513 & 0.2434 \\
     & ATS+SAT(GCN) & 0.0426 & 0.0765 & 0.1631 & 0.1039 & 0.1491 & 0.2411 \\
     & ATS+SAT(GAT) & \textbf{0.0438} & \textbf{0.0785} & \textbf{0.1651} & \textbf{0.1067} & \textbf{0.1529} & \textbf{0.2450} \\
    \bottomrule
    \end{tabular}
}}
\end{table*}

\subsubsection{Profiling}
The model outputs the restored attributes in different dimensions with probabilities.
Higher corresponding probabilities of ground-truth attributes signify better performance. In this section, we use two common metrics Recall@k and NDCG@k to evaluate the profiling performance between new model and original model. The experiment results are shown in Table~\ref{tab:profiling}.

According to the profiling results in Table~\ref{tab:profiling}, the majority of models combined with ATS method perform better than the original models. On the basis of the advantages established by the SAT model towards other baselines, the combination of the ATS algorithm and SAT model (ATS+SAT) obtains even higher performance in almost all the evaluation metrics and almost all the datasets. For example, ATS+SAT(GAT) obtains a relative 13.1\% gain of Recall@10 and a relative 12.5\% gain of NDCG@10 on Citeseer compared with SAT(GAT). Furthermore, ATS method can also augment other base model's profiling performance. ATS+GAT achieves a relative 17.4\% higher performance of Recall@10 and a relative 24.9\% higher performace of NDCG@10 on Citeseer compared with original GAT.
The main reason of these results is that the active sampling algorithm ATS helps these base models realize different importance of different nodes in learning, and achieves more accurate distribution modeling of the high-dimensional node attributes.

\subsection{Ablation Study}
\subsubsection{Effect of Different Node Degree}

ATS will preferentially select the nodes with larger amount of information to make primary model extract more information from more informative nodes. We thus design an experiment to verify whether ATS can facilitate the learning at different information levels and how the information level influences the performance. In particular, Node degree is an important indicator for centrality metric, so we sort the nodes in the test set according to their degrees and select a ratio of nodes with a range of degrees for experiment. For example, every 20\% of test nodes sorted by degrees corresponds to one Degree Level. The results are shown in Table~\ref{tab:degrees}.

\begin{table*}[!htp]
\captionsetup{font=small}
    \caption{Recall@20 results for profiling task on nodes with different degrees. Test set is divided into 5 equal parts according to the degree value, and each subset represents one Degree Level. Higher Degree Level indicates larger node degree.}
     \label{tab:degrees}
    \centering
    \normalsize
\renewcommand{\arraystretch}{0.7}
 \setlength{\tabcolsep}{1.5mm}{ 
  \scalebox{0.75}{
    \begin{tabular}[c]{c|ccccc|ccccc}
    \toprule
      & \multicolumn{5}{c}{Citeseer} & \multicolumn{5}{c}{Amazon-Computer} \\ \hline
    Degree Level & 1 & 2 & 3 & 4 & 5 & 1 & 2 & 3 & 4 & 5 \\
    \midrule

         SAT(GCN) & 0.1183 & 0.1072 & 0.1252 & 0.1400 & 0.1673 & 0.0603 & 
0.0728 & 0.0740 & 0.0758 & 0.0774 \\
        ATS+SAT(GCN) & 0.1274 & 0.1103 & 0.1343 & 0.1479 & 0.1796 & 0.0621 & 0.0750 & 0.0770 & 0.0789 & 0.0800 \\
        Improvement(\%) & 0.91 & 0.31 & 0.91 & 0.79 & \textbf{1.23} & 0.18 & 
0.22 & 0.30 & \textbf{0.31} & 0.26 \\ \hline
& \multicolumn{5}{c}{Citeseer} & \multicolumn{5}{c}{Amazon-Computer} \\ \hline
         SAT(GAT) & 0.1255 & 0.1130 & 0.1326 & 0.1429 & 0.1739 & 0.0610 & 0.0761 & 0.0763 &  
0.0787 & 0.0818 \\ 
         ATS+SAT(GAT) & 0.1329 & 0.1193 & 0.1425 & 0.1602 & 0.1873 & 0.0604 & 0.0756 & 0.0768 & 0.0789 & 0.0828 \\
         Improvement(\%) & 0.74 & 0.63 & 0.99 & \textbf{1.73} & 1.34 & -0.06 & -0.05 & 0.05 & 0.02 & \textbf{0.1} \\
    \bottomrule
    \end{tabular}
}}
\end{table*}

On one hand, ATS can achieve improvements at almost all degree levels compared to the original primary model. Even less informative nodes benefit from our ATS's training strategy. The reason is that ATS is capable of capturing the different importance of nodes in the learning schedule, helping the primary model converge to a better state. On the other hand, when considering the improvement gap along with degree levels, we find that the gap generally becomes more evident as the node degree level increases. This result verifies the correctness of ATS's idea that employs node information in importance modeling.

\subsubsection{Different Weighting Scheme}

Besides the active sampling metrics, the Beta distribution controlled weighting scheme is also a highlight of the ATS algorithm. We will verify the effectiveness of our proposed scheme in comparison with other weighting schemes, such as the fixed weighting scheme and the linear variation weighting scheme. For the fixed one, the values of $\gamma$ are $0.2, \frac{1}{3}, 0.6$, and $\alpha = \beta = \frac{1 - \gamma}{2}$. For the linear variation one, $\gamma$ decreases linearly from 1 to 0.5 or from 1 to 0.

\begin{figure}[!htp]
\centering
\subfigure[Cora]{\begin{minipage}[t]{0.32\textwidth}
\centering
\includegraphics[width=\textwidth]{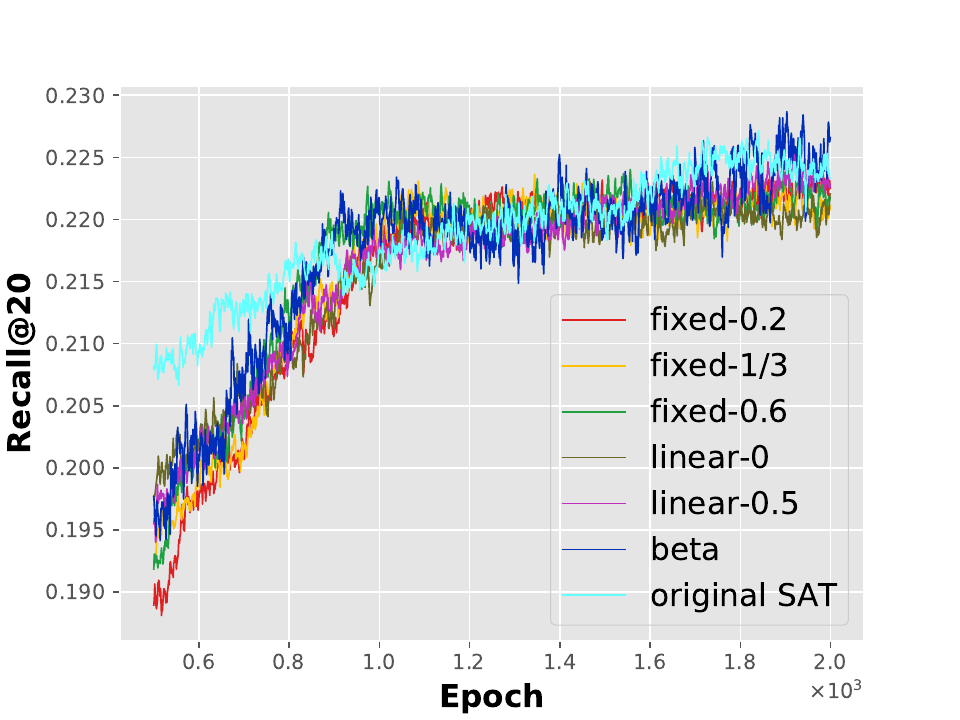}
\end{minipage}}
\subfigure[Citeseer]{\begin{minipage}[t]{0.32\textwidth}
\centering
\includegraphics[width=\textwidth]{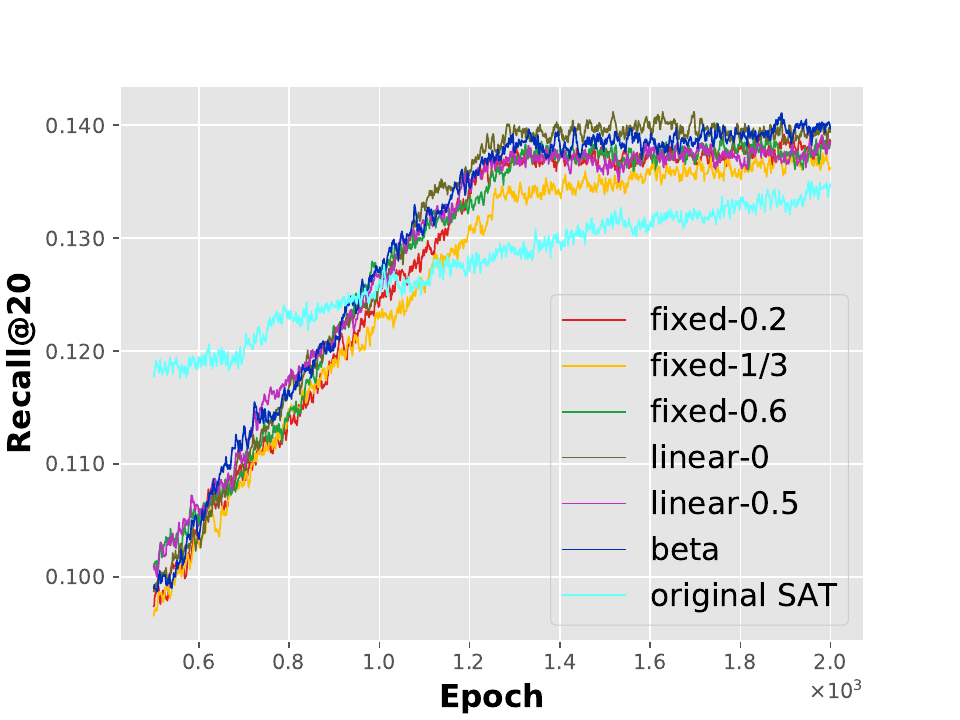}
\end{minipage}}
\subfigure[Amazon-Photo]{\begin{minipage}[t]{0.32\textwidth}
\centering
\includegraphics[width=\textwidth]{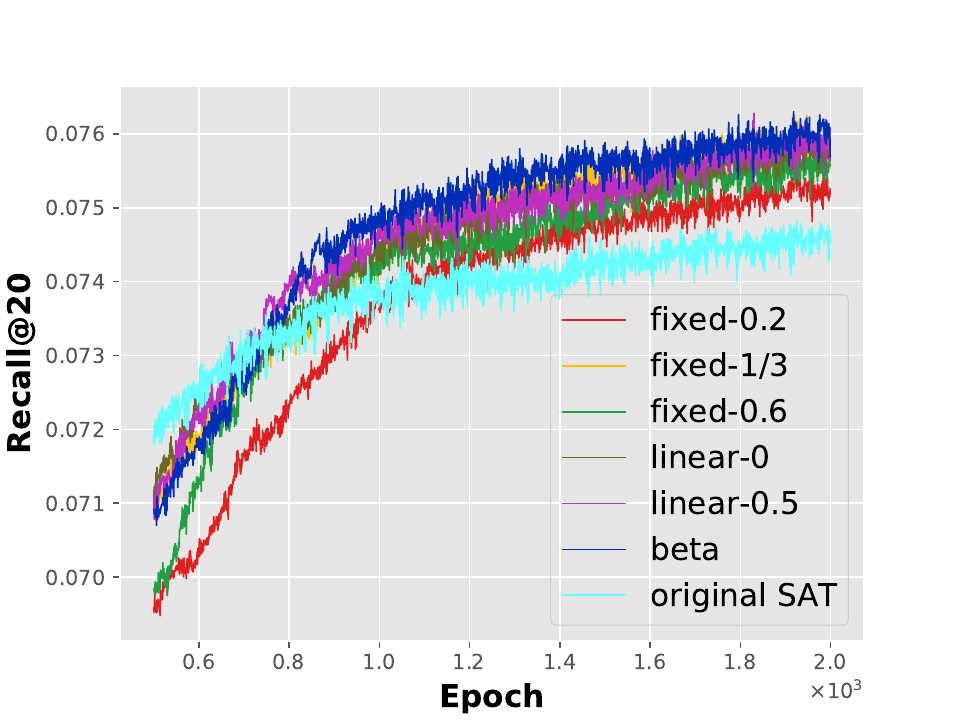}
\end{minipage}}
\captionsetup{font=small}
\caption{Visualization of profiling performance of different weighting schemes on test data during training process. We compare our Beta distribution controlled weighting scheme with other weighting schemes(e.g. fixed weight, linear variation) and also the original SAT model.}
	\label{fig: scheme}
\end{figure}

From Figure \ref{fig: scheme}, our proposed weighting scheme outperforms other schemes and original SAT model because Beta distribution changes the weights dynamically during the sampling process and meanwhile remains some randomness to improve the robustness of the algorithm. The results on three datasets show that our method converges faster than the uniform sampling of original SAT and other schemes like fixed or linear weighting. More importantly, our method converges to a better state with higher performance.

\subsubsection{Different Centrality Metrics}

In Section~\ref{sec: query strategy}, we mention that structural centrality is evaluated using the PageRank score. Meanwhile, there are several centrality metrics other than PageRank including degree centrality, closeness centrality, betweenness centrality and eigenvector centrality. All of these metrics can reflect the structural information contained in each node. We will compare the performance of these different metrics using the SAT(GCN) primary model. All these five metrics are implemented using NetworkX\footnote{https://networkx.github.io/}.

\begin{figure*}[!htp]
\centering
\subfigure[Cora]{\begin{minipage}[t]{0.4\textwidth}
\centering
\includegraphics[width=\textwidth]{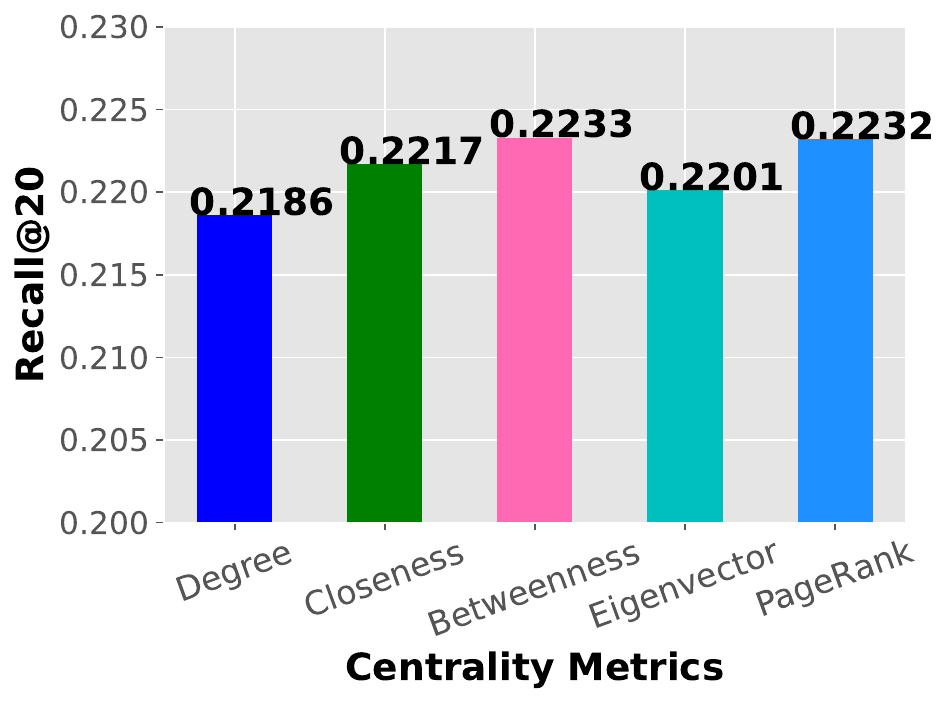}
\end{minipage}}
\subfigure[Citeseer]{\begin{minipage}[t]{0.4\textwidth}
\centering
\includegraphics[width=\textwidth]{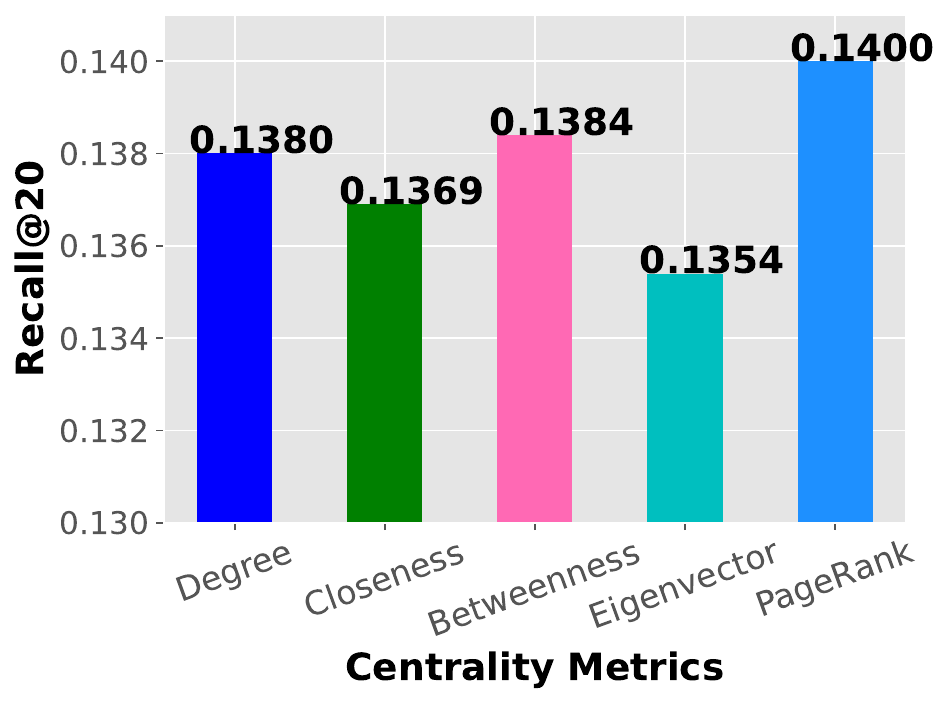}
\end{minipage}}
\subfigure[Amazon-Photo]{\begin{minipage}[t]{0.4\textwidth}
\centering
\includegraphics[width=\textwidth]{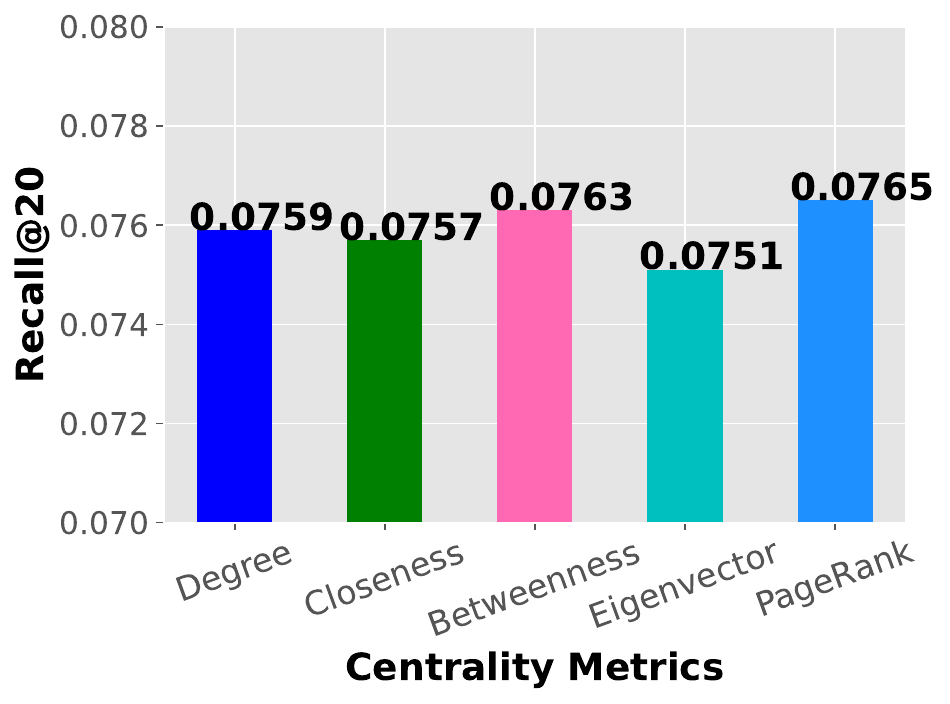}
\end{minipage}}
\subfigure[Amazon-Computer]{\begin{minipage}[t]{0.4\textwidth}
\centering
\includegraphics[width=\textwidth]{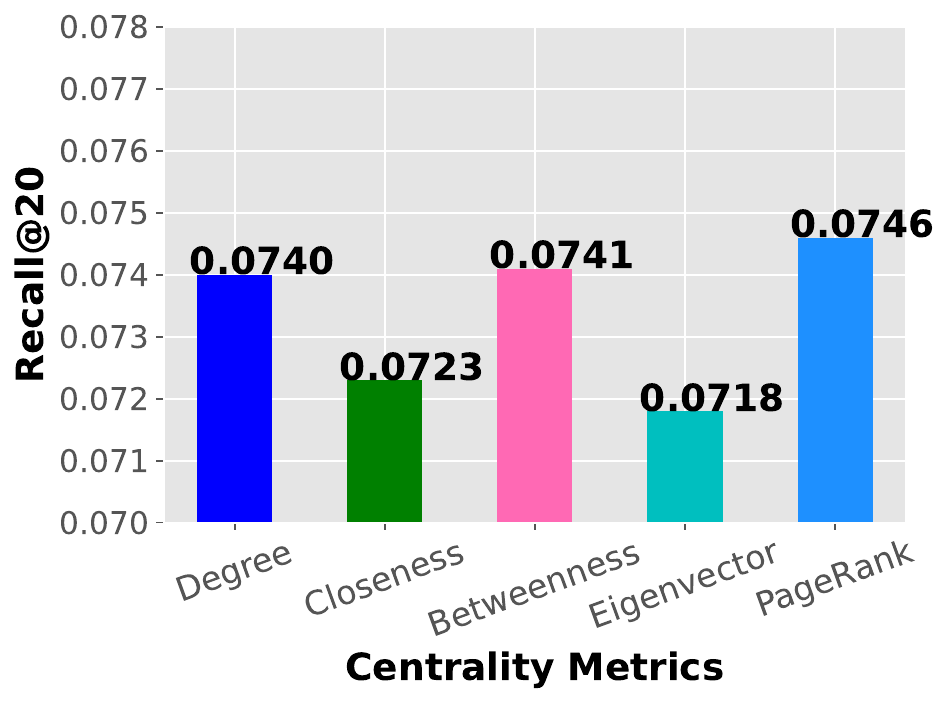}
\end{minipage}}
\captionsetup{font=small}
\caption{Comparison among Recall@20 performance using different centrality metrics on four datasets. The primary model is SAT(GCN).}
\label{fig: centrality}
\end{figure*}

The result is shown in Figure~\ref{fig: centrality}. The PageRank metric outperforms other metrics on almost all the datasets, indicating that PageRank can better leverage structural information. Therefore, choosing the PageRank score rather than other metrics as the structural centrality is reasonable.

\subsubsection{Different Metric Combinations}
\begin{figure}[!htp]
\centering
\subfigure[Cora]{\begin{minipage}[t]{0.3\textwidth}
\centering
\includegraphics[width=\textwidth]{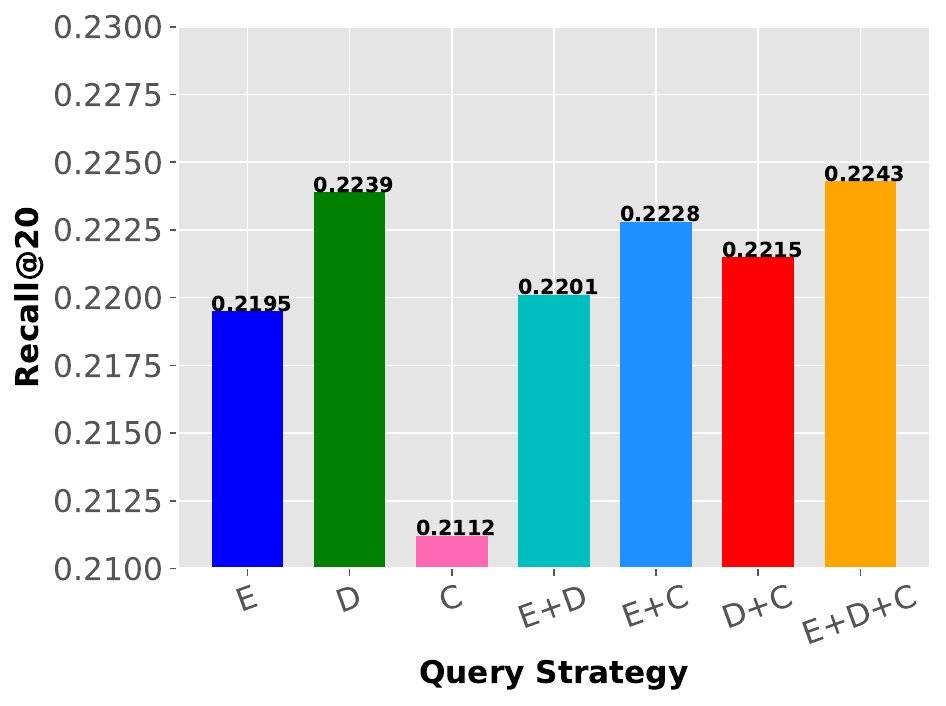}
\end{minipage}}
\subfigure[Citeseer]{\begin{minipage}[t]{0.3\textwidth}
\centering
\includegraphics[width=\textwidth]{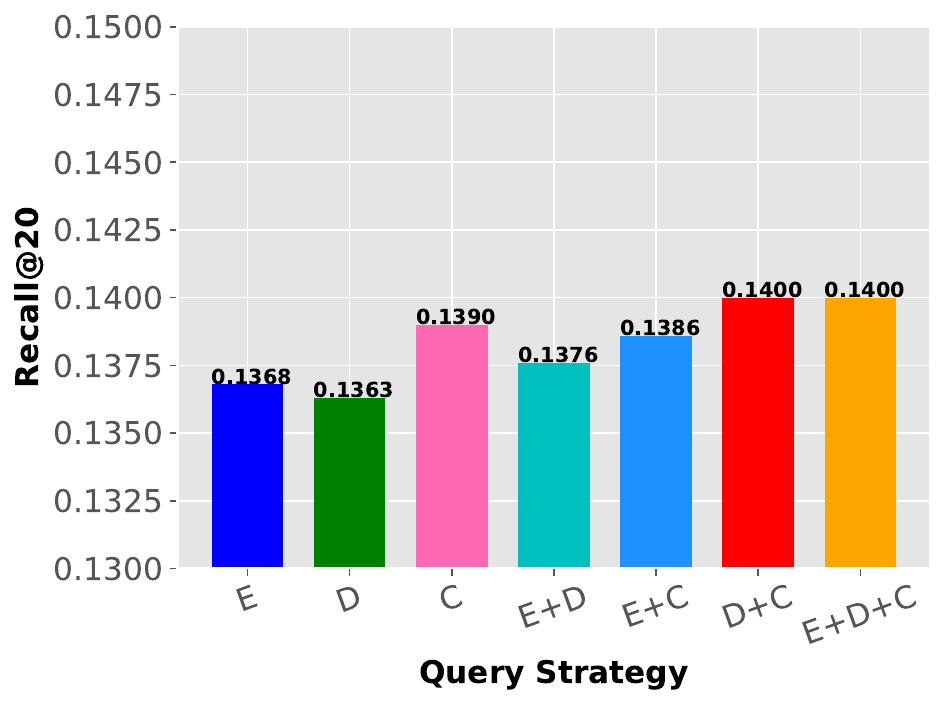}
\end{minipage}}
\subfigure[Amazon-Photo]{\begin{minipage}[t]{0.3\textwidth}
\centering
\includegraphics[width=\textwidth]{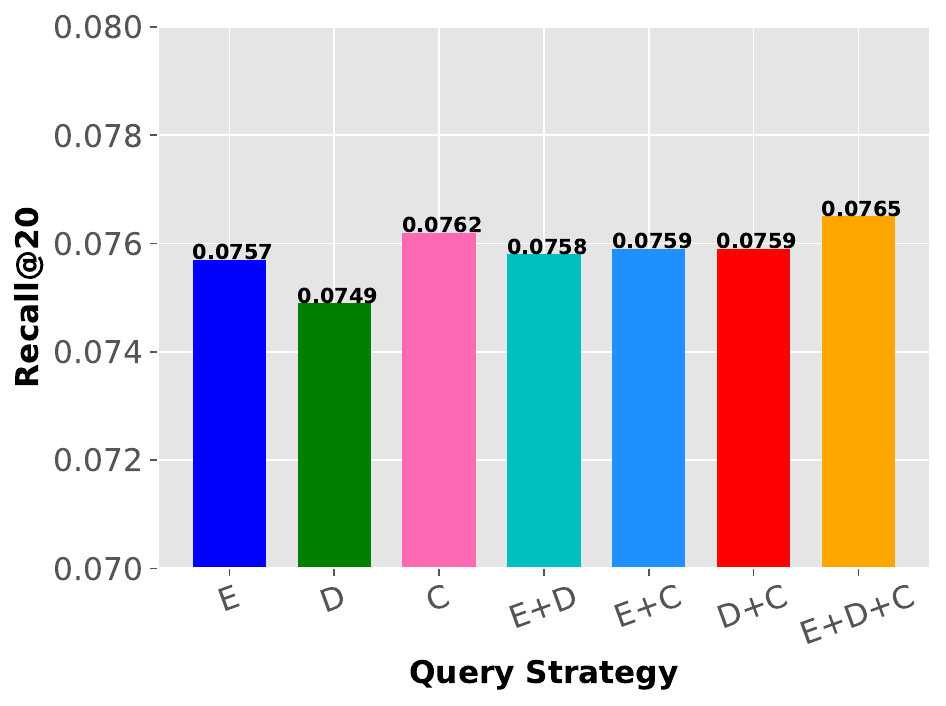}
\end{minipage} }

\captionsetup{font=small}
\caption{Ablation study of different metrics in ATS. We show the recall@20 result of different combinations of the sampling metrics on 3 benchmarks. The horizontal coordinate refers to the different sampling criteria combinations. 'E' indicates the entropy metric; 'D' indicates the density metric; 'C' indicates the centrality metric; 'E+D+C' indicates our ATS algorithm.}
\label{fig: ablations}
\end{figure}

In this section, we conduct the ablation study to investigate the effects of three different metrics in ATS.
The experimental settings remain the same as the profiling task. We use Recall@20 to evaluate the performance of different metric combinations. The results are shown in Figure~\ref{fig: ablations}.

On all datasets, incomplete sampling metrics of ATS would achieve inferior performance.
Different metrics focus on different aspects and the result shows that they can complement each other. The uncertainty metric focuses on the training status of the model, while the representativeness metric focuses on the implied information from both the structure and attribute aspects.
Combining them all can help the node attribute completion model better capture the importance of the node in learning and achieve better performance.

\subsection{Empirical Time Complexity Analysis}

\begin{figure*}[!htp]
\centering
\subfigure[Cora]{\begin{minipage}[t]{0.4\textwidth}
\centering
\includegraphics[width=\textwidth]{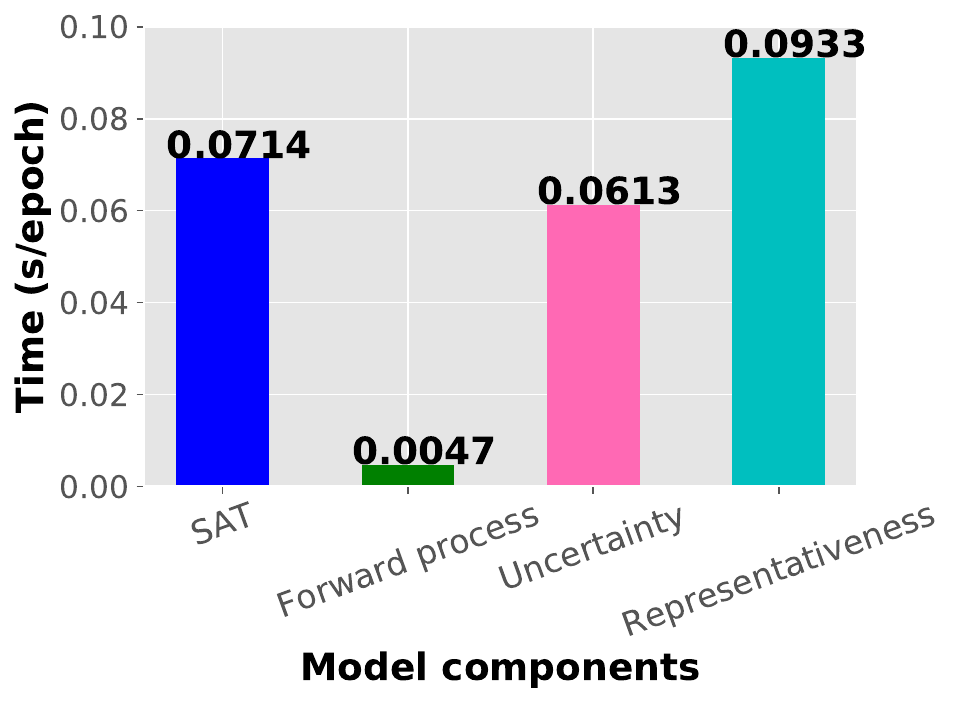}
\end{minipage}}
\subfigure[Citeseer]{\begin{minipage}[t]{0.4\textwidth}
\centering
\includegraphics[width=\textwidth]{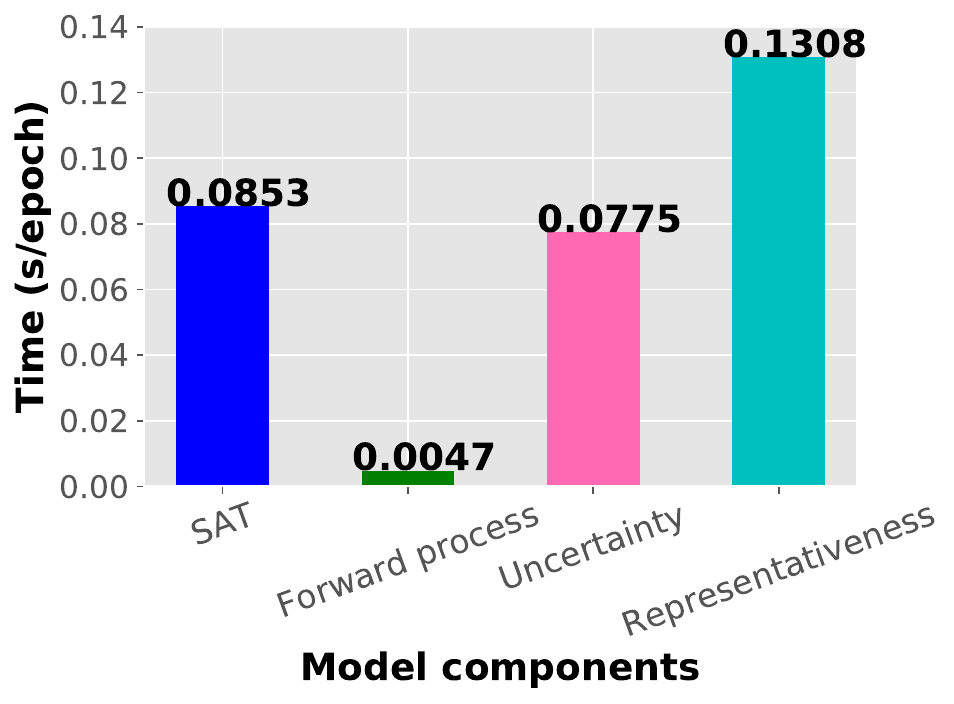}
\end{minipage}}
\subfigure[Amazon-Photo]{\begin{minipage}[t]{0.4\textwidth}
\centering
\includegraphics[width=\textwidth]{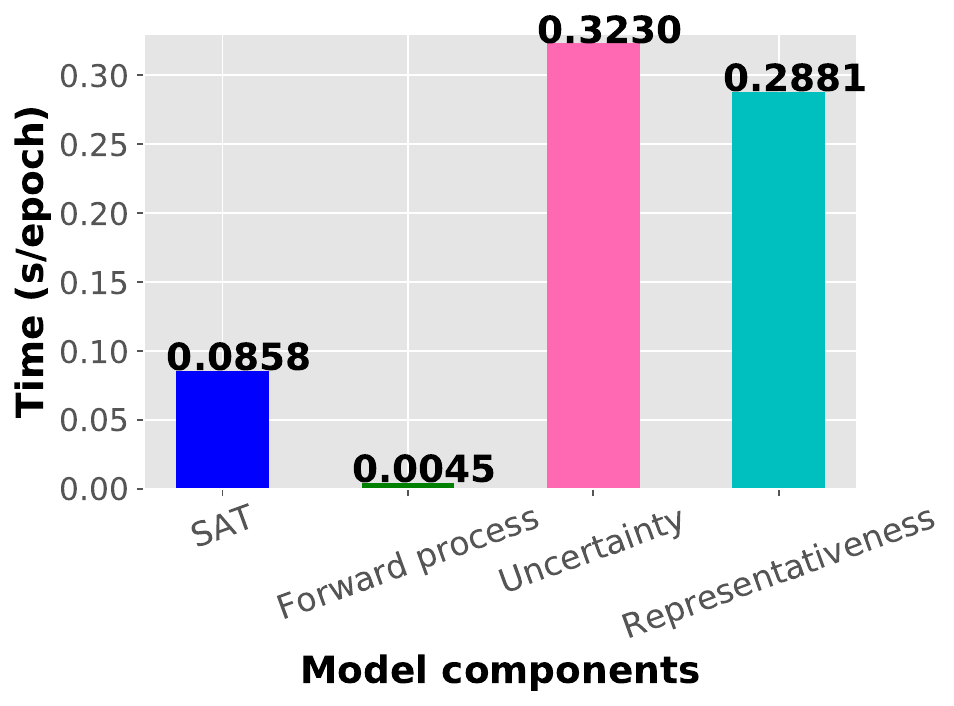}
\end{minipage}}
\subfigure[Amazon-Computer]{\begin{minipage}[t]{0.4\textwidth}
\centering
\includegraphics[width=\textwidth]{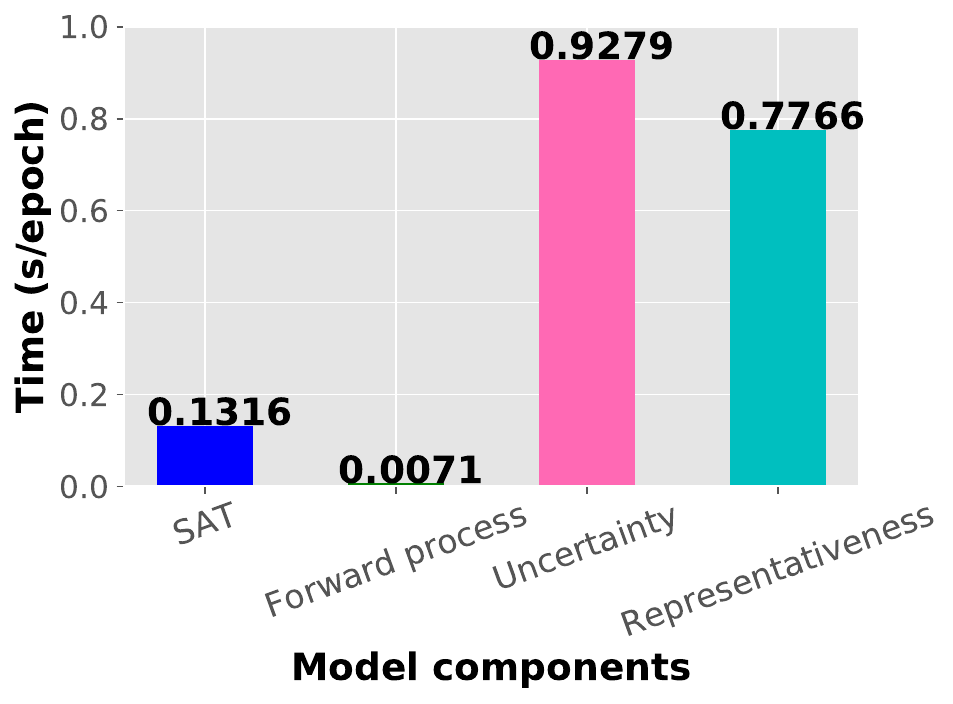}
\end{minipage}}
\captionsetup{font=small}
\caption{The comparison among the average processing GPU time per epoch of different components on different datasets. 'Forward' indicates the forward propagation that is part of the calculation in the uncertainty metric.}
\label{fig: time}
\end{figure*}

Our ATS is an active sampling procedure based on a node attribute completion model, so it is critical to study the extra processing time cost by ATS. Thus, we conduct an experiment to count the running time of different parts of ATS compared with the original primary base model--SAT. These different parts are forward process, uncertainty and representativeness. The forward process means the forward propagation, which is essential to calculate the uncertainty score. We implemented the experiment on a machine with one Nvidia 1080Ti GPU.
According to the running time shown in Figure \ref{fig: time}, forward propagation in ATS takes only a small part of time in SAT since back propagation usually costs a lot of time. 
Although the processing time of uncertainty metric and representativeness metric is relatively higher than SAT because of the clustering and percentile calculations, it's comparable with the time of SAT. With the addition of the ATS algorithm, the time required for each epoch will increase within an acceptable range.

\section{Conclusion and Future Work}
\vspace{-3pt}
In this paper, we propose a novel active sampling algorithm ATS to better solve the problem of node attribute completion. The ATS can also be combined with other primary base models that have latent representations and training loss depending on different tasks. In order to distinguish the differences in the amount of information among nodes, ATS utilizes the proposed uncertainty and representativeness metrics to select the most informative nodes and renew the training set after each training epoch. In addition, the Beta distribution controlled weighting scheme is proposed to dynamically adjust the metric weights according to the training status. The sampling process increases the running time of each epoch within an affordable cost but meanwhile helps the primary base model achieve superior performance on profiling and node classification tasks. Therefore, ATS is effective in boosting the quality of restored attributes and is applicable to different primary models.
In future, we will extend ATS to more primary models and applications.

\bibliographystyle{elsarticle-num} 
\bibliography{reference.bib}

\appendix
\section{Details About the Baselines}
\label{appendix:baseline_details}
NeighAggre is an intuitive attribute aggregation algorithm. It completes one node's missing attributes by averaging its neighbour nodes' attributes, which is a simple but efficient method to take advantage of the structural information. VAE is a famous generative model that consists of an encoder and a decoder. For test nodes without the attributes, the encoder will generate the corresponding latent code through the neighbour aggregation. Then the decoder will restore the missing attributes. GCN, GraphSage and GAT are three typical graph representation learning methods. For attribute-missing scenario, only the graph structure will be encoded to latent codes. The missing attributes will be recovered by the decoders of these GNN methods from the latent code generated by the encoders. Hers is a cold-start recommendation method. GraphRNA and ARWMF are two attributed random walk based methods to learn the node representations, which can be extended to
deal with the missing attribute problems. They separate the graph structure and node attributes and learn the node embeddings by random walks. PaGNNs are  novel message propagation schemes for attribute-incomplete graphs. They leverage a deeper representation of the nodes and perform well on node classification task.

\subsection{Sensitivity of the Hyperparameters}
As mentioned in Section 5.2, cluster number is a vital hyper-parameter that determines the information density of each node.
We conduct the experiments on both the profiling and classification tasks with different cluster numbers.

\begin{figure}[!htp]
	\centering
	\subfigure[Recall on Cora]{
		\begin{minipage}[t]{0.3\linewidth}
			\centering
			\includegraphics[width=3cm]{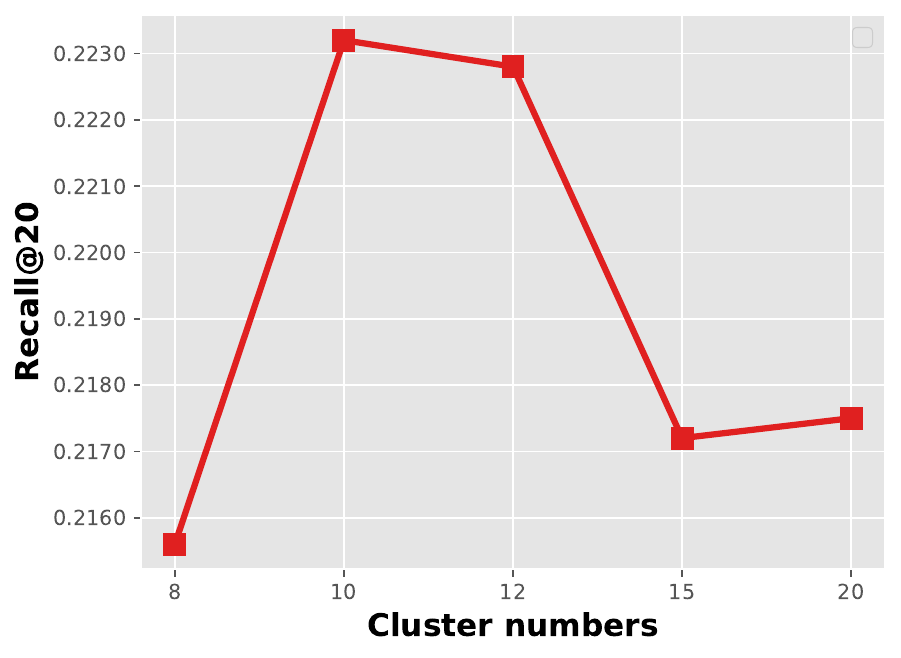}
		\end{minipage}
	}%
	\hspace{0.03\linewidth}
	\subfigure[Recall on Citeseer]{
		\begin{minipage}[t]{0.3\linewidth}
			\centering
			\includegraphics[width=3cm]{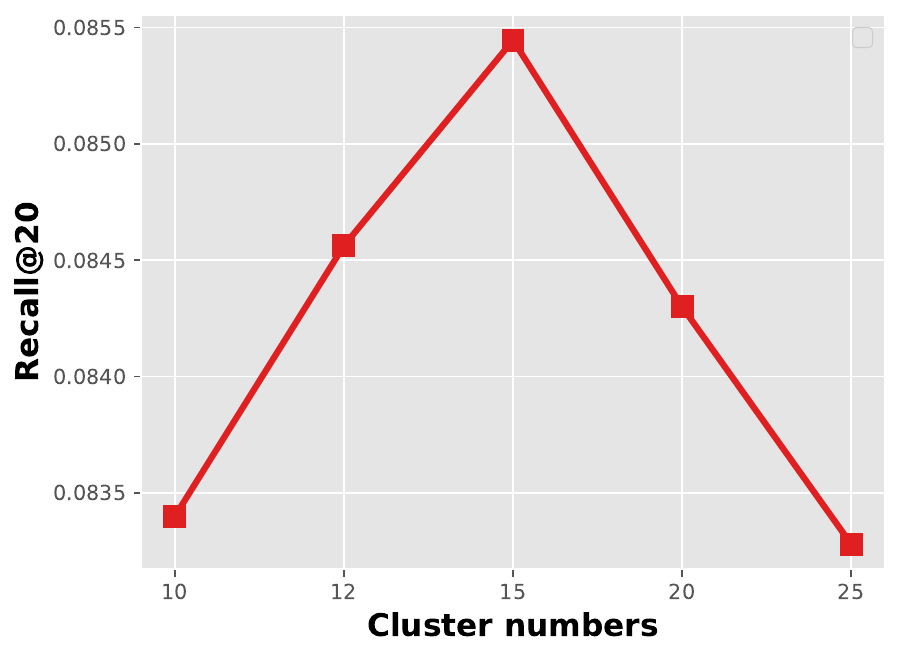}
		\end{minipage}
	}%
	\hspace{0.03\linewidth}
	\subfigure[Recall on Ama\_Pho]{
		\begin{minipage}[t]{0.25\linewidth}
			\centering
			\includegraphics[width=3cm]{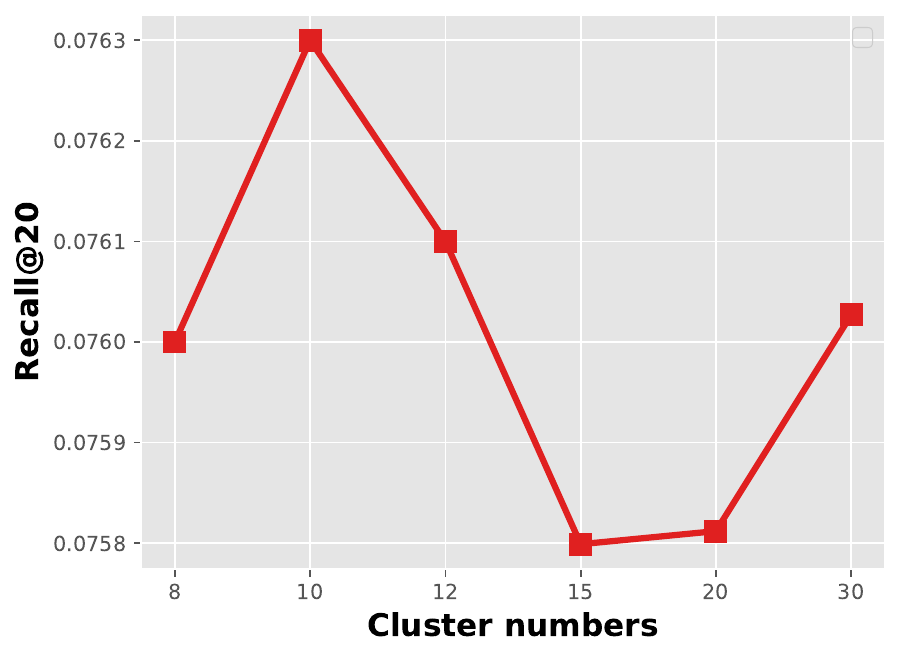}
		\end{minipage}
	}%
	
	\subfigure[X - Cora]{
		\begin{minipage}[t]{0.3\linewidth}
			\centering
			\includegraphics[width=3cm]{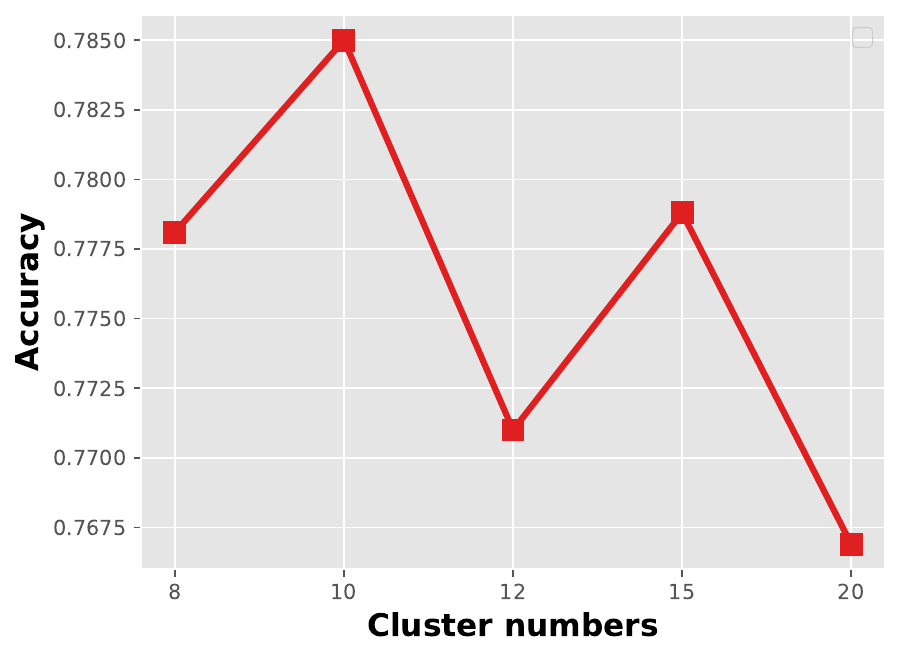}
		\end{minipage}
	}%
	\hspace{0.03\linewidth}
	\subfigure[X - Citeseer]{
		\begin{minipage}[t]{0.3\linewidth}
			\centering
			\includegraphics[width=3cm]{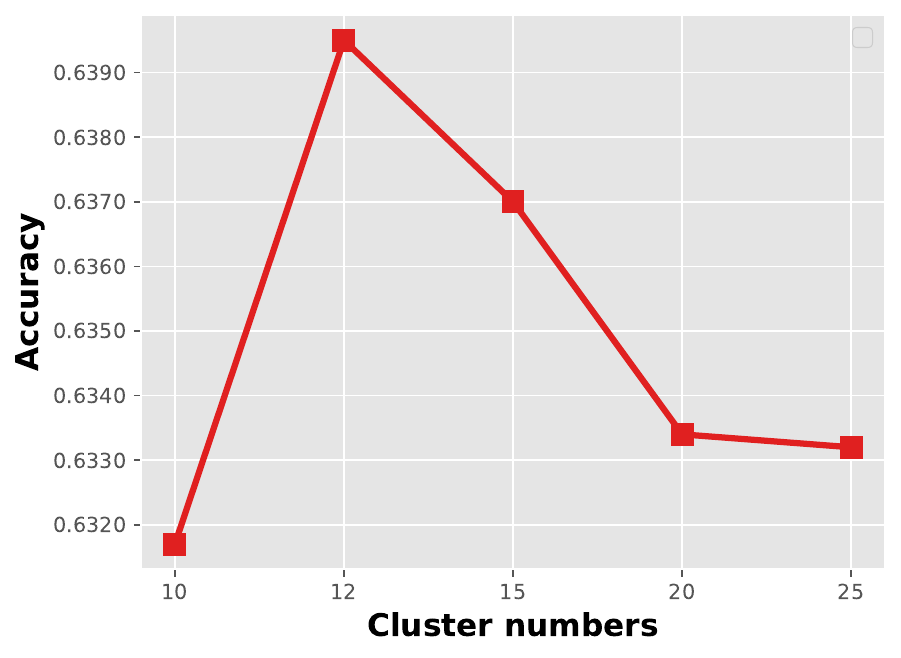}
		\end{minipage}
	}%
	\hspace{0.03\linewidth}
	\subfigure[X - Ama\_Pho]{
		\begin{minipage}[t]{0.25\linewidth}
			\centering
			\includegraphics[width=3cm]{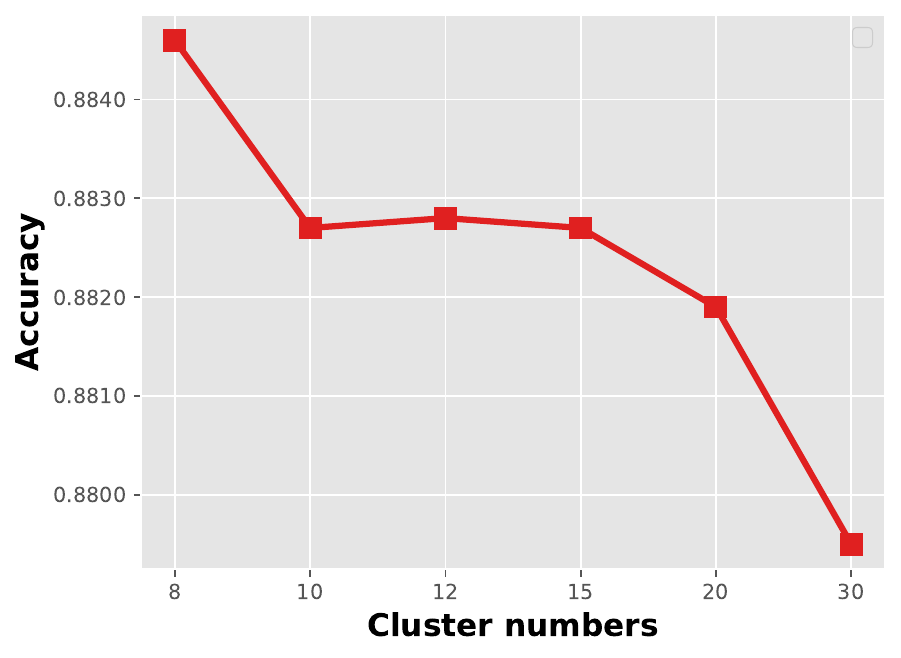}
		\end{minipage}
	}%
	
	\subfigure[A+X - Cora]{
		\begin{minipage}[t]{0.3\linewidth}
			\centering
			\includegraphics[width=3cm]{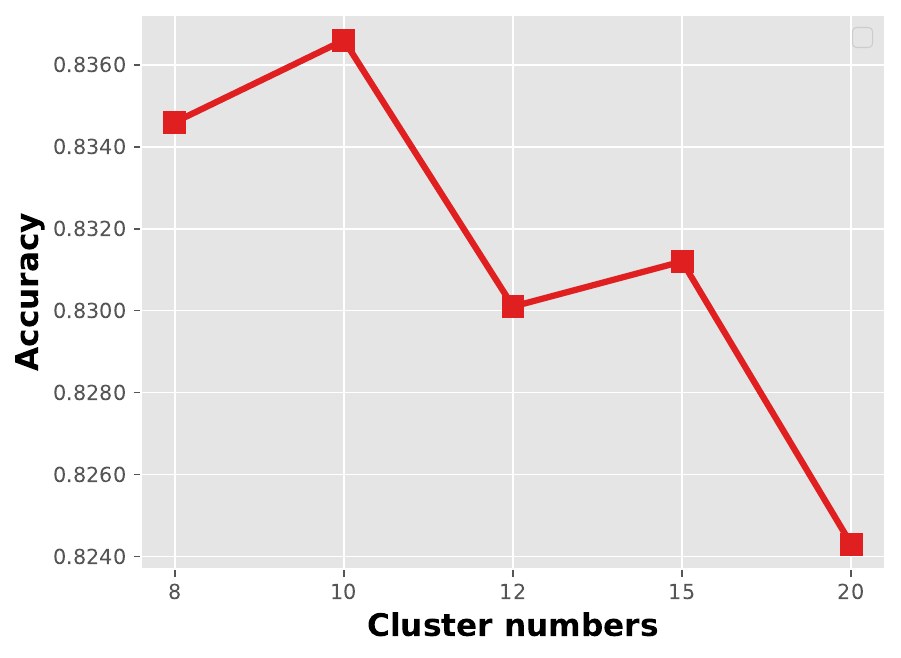}
		\end{minipage}
	}%
	\hspace{0.03\linewidth}
	\subfigure[A+X - Citeseer]{
		\begin{minipage}[t]{0.3\linewidth}
			\centering
			\includegraphics[width=3cm]{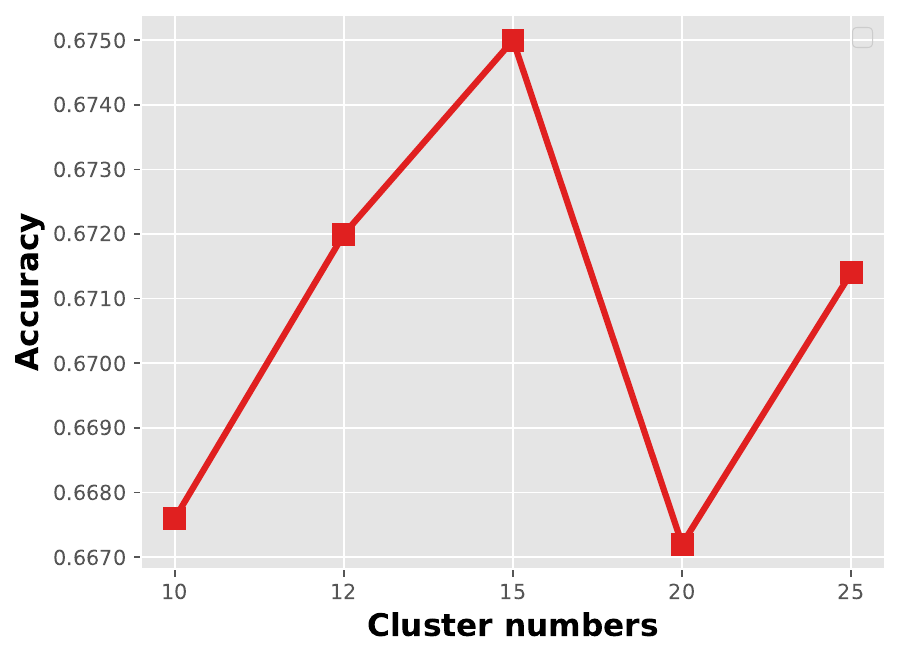}
		\end{minipage}
	}%
	\hspace{0.03\linewidth}
	\subfigure[A+X - Ama\_Pho]{
		\begin{minipage}[t]{0.25\linewidth}
			\centering
			\includegraphics[width=3cm]{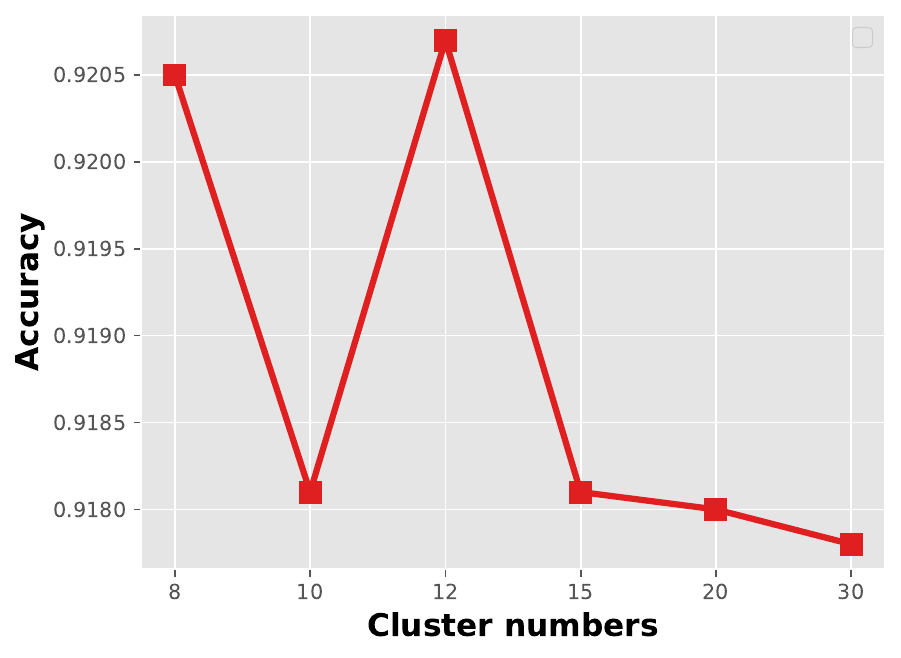}
		\end{minipage}
	}%
	\centering
	\caption{Results with different cluster numbers when calculating the density score in the representativeness metric. (a-c) show the Recall@20 results for profiling task. (d-f) show the attribute-only classification accuracy with the use of MLP classifier. (g-h) show the classification accuracy considering both the structure and attribute information.}
	\label{fig: hyper}
\end{figure}

The results of Figure \ref{fig: hyper} show that too large or too small cluster numbers are not conducive to the training. If there are not enough cluster centers, the sampling algorithm is not robust to extract the density of the embedding distribution. On the other hand, if there are too many cluster centers, it will introduce more disturbance and might separate the nodes belonging to the same class. We implement grid-search to determine the value of hyperparameter based on the Recall@20 results in the profiling task.









\end{document}